\documentclass[10pt, conference]{IEEEtran}

\usepackage{cite}
\usepackage[pdftex]{graphicx}
\usepackage[cmex10]{amsmath}
\usepackage{comment}

\usepackage{algorithm}
\usepackage{algorithmic}

\usepackage{tabularx}
\usepackage{multicol}

\usepackage{boldline}

\usepackage{subfig}

\usepackage{mathtools}
\DeclarePairedDelimiter{\ceil}{\lceil}{\rceil}

\makeatletter
\def\footnoterule{\relax%
  \kern-5pt
  \hbox to \columnwidth{\hfill\vrule width 1\columnwidth height 0.4pt\hfill}
  \kern4.6pt}
\makeatother

\usepackage{textcomp}
\def\BibTeX{{\rm B\kern-.05em{\sc i\kern-.025em b}\kern-.08em
		T\kern-.1667em\lower.7ex\hbox{E}\kern-.125emX}}

\graphicspath{{figs/}}

\usepackage{amsmath}
\newcommand{\inlineeqnum}{\refstepcounter{equation}~~\mbox{(\theequation)}}

 \usepackage{background}
 \backgroundsetup{angle=0,  
   scale=1, 
   color=black, 
   firstpage=true,
   position=current page.north, 
   hshift=0pt,
   vshift=-20pt,
   contents={\ifnum\value{page}=1 PREPRINT: Accepted at 28th International Conference on Field Programmable Logic \& Applications (FPL), 2018 \else \fi}
 }

\begin{document}

\title{$\mathbf{Cascade^{C_{N_N}}}$: Pushing the Performance Limits of Quantisation in Convolutional Neural Networks \vspace{-0.35cm}}

\author{
\IEEEauthorblockN{Alexandros Kouris, Stylianos I. Venieris and Christos-Savvas Bouganis}
\IEEEauthorblockA{Dept. of Electrical and Electronic Engineering, Imperial College London\\ Email: \{a.kouris16, stylianos.venieris10, christos-savvas.bouganis\}@imperial.ac.uk}
\vspace{-0.92cm}
}

\maketitle

\begin{abstract}
This work presents \textit{CascadeCNN}, an automated toolflow that pushes the quantisation limits of any given CNN model, aiming to perform high-throughput inference. 
A two-stage architecture tailored for any given CNN-FPGA pair is generated, consisting of a low- and high-precision unit in a cascade. A confidence evaluation unit 
is employed to identify misclassified cases from the excessively low-precision unit and forward them to the high-precision unit for re-processing. Experiments demonstrate that the proposed toolflow can achieve a performance boost up to 55\% for \mbox{VGG-16} and 48\% for AlexNet over the baseline design for the same resource budget and accuracy, without the need of retraining the model or accessing the training data.
\end{abstract}


\IEEEpeerreviewmaketitle




\vspace{-2mm}
\section{Introduction}
\vspace{-1mm}
While Convolutional Neural Networks (CNNs) are becoming the state-of-the-art algorithm in various Machine Vision tasks, 
they are challenged to deal with problems of continuously increasing complexity. The significant advances of CNNs came with increased number of layers \cite{simonyan2014very}, increased number of kernels \cite{zeiler2014visualizing} and more complex architectures \cite{he2016deep}, which introduce substantial 
costs in terms of computational and memory resources. 
To deploy CNNs in real-world tasks which deal with vast amounts of data, it is necessary that the high computation and memory requirements of such models are alleviated. 
To this end, numerous compression and precision quantisation techniques have been proposed \cite{hubara2016quantized,han2015deep,wu2016quantized} which exploit the redundancy in CNN models to enable their efficient deployment on processing platforms.

In this context, FPGAs constitute a promising platform for 
CNN inference due to their customisability which enables the use of optimised low-precision arithmetic units to achieve performance gains \cite{sv2018csur}. Existing FPGA-based accelerators have produced hardware designs that span from uniform \mbox{16-bit} precision \cite{Venieris_2016}\cite{Yufei_Ma_2017} 
with minimal effect on accuracy, down to very high-performance binarised networks \cite{Umuroglu_2017}, but at a significant accuracy loss. In this setting, given a fixed resource budget, the attainable performance for a given error tolerance is limited by the shortest wordlength that meets the error bound.

In this paper, we propose \textit{CascadeCNN}, a novel automated methodology for generating a high-throughput cascade of CNN classifiers that pushes the performance of precision quantised CNNs. \textit{CascadeCNN} exploits the fact that not all inputs require the same level of precision to obtain a confident prediction and introduces precision quantisation as a design dimension for the deployment of CNN cascades. In this respect, the proposed framework generates an excessively low-precision processing unit to obtain rapid classification prediction together with a parametrised mechanism for estimating the prediction confidence. The samples that are detected as misclassified are recomputed on a high-precision unit to restore the application-level accuracy and meet the user-specified limits. \textit{CascadeCNN} considers the error tolerance and the target CNN-device pair to select quantisation scheme, configure the confidence evaluation mechanism and generate the cascaded low- and high-precision processing units. 
The main contributions of this work are the following:
\begin{itemize}
    \item The introduction of the first CNN cascade in the literature that exploits the custom arithmetic potential of FPGAs to boost CNN inference throughput. A novel tunable confidence evaluation unit is presented which estimates the confidence of classification predictions at run time. By considering the user-specified error tolerance for the target application, the confidence evaluation unit decides accordingly whether or not an input sample is to be recomputed by the second stage of the cascade.
    
    \item A novel FPGA-based architecture which scales its performance as the wordlength of activations and weights decreases, by fully exploiting the available FPGA resources. The proposed architecture employs both DSP blocks and LUTs to map its compute units and demonstrates performance gains with each bit reduction. The architecture is able to execute both the convolutional and fully-connected layers of CNNs and is parametrised so that it can be configured at compile time.
    
    \item A design space exploration methodology that takes into account the CNN-FPGA pair and a user-defined error tolerance for the target application and optimises the CNN cascade structure with respect to the quantisation scheme and the hardware configuration for each stage. Furthermore, this work employs the full reconfiguration of the FPGA as a design option, to enable the proposed cascade scheme on FPGAs whose resource budget does not allow both stages to be instantiated at once, along with appropriate batch-processing to maintain high throughput by alleviating the performance cost of reconfiguration.
\end{itemize}


\vspace{-2mm}
\section{Background and Related Work}
\vspace{-1mm}

\subsection{Cascade Machine Learning Systems}
\vspace{-1mm}
Cascades of classifiers is a widely studied design approach in machine learning \cite{Viola2004robust}\cite{Xu2014jmlrcascade}. Such algorithms aim to minimise the computation time per classification by exploiting the property that different inputs require different amount of computation to obtain a confident prediction. A cascade structure is typically formed by connecting classifiers of increasing complexity as a multi-stage architecture. At each stage, based on the confidence of its prediction, a classifier can either decide upon the classification of the current input sample and terminate the execution or pass it to the next stage. The deep learning community has mainly focused on the design of CNN cascades from an algorithmic perspective targeting diverse tasks including the detection of faces \cite{Li2015cnncascade}, pedestrians \cite{Angelova2015cnncascade} and objects \cite{Diba2017cnncascade}. In a more systems-oriented approach, Kang et al. \cite{Kang2017noscope} proposed \textsc{NoScope}, an automated framework that generates CNN cascades optimised for high-throughput query-based video analysis on high-end GPUs, by searching the design space of both CNN models and cascade structures with the objective to perform binary classification tasks.


In contrast to the existing literature, this work approaches CNN cascade design from a different perspective. Instead of designing one weaker and one more sophisticated model at the training phase, \textit{CascadeCNN} takes as input a model trained at a given precision and generates one weaker but faster, low-precision model and one slower but more accurate, high-precision model. To decide whether to pass each sample to the second-stage classifier, a novel tunable confidence evaluation unit is introduced, which estimates the classification confidence of the low-precision model. By taking into account the user-specified error tolerance, the two stages of the cascade, the corresponding hardware architectures and the confidence evaluation unit are co-optimised to maximise performance while lying within the acceptable user-defined error bounds. Please note that this precision-oriented cascade approach is orthogonal to model cascades and can be applied independently to each of their stages to improve their performance. 


\vspace{-3mm}
\subsection{Precision Quantisation of CNN Models}
\vspace{-1.5mm}
Precision quantisation is an actively researched method for minimising the computational and space complexity of CNN inference. 
The majority of existing works apply fixed-point quantisation to full-precision trained models and employ a retraining step as a mechanism of fine-tuning the network's fixed-point weights. 
In this direction, Gysel et al. \cite{Gysel2016ristretto} proposed Ristretto, a framework that quantises both activations and weights under a dynamic fixed-point scheme with uniform wordlength and different scaling across layers. 
\mbox{Angel-Eye \cite{Guo2017angeleye}} is a CNN-to-FPGA framework that comprises a quantisation method together with a CNN accelerator. Similarly to Ristretto, Angel-Eye employs uniform wordlength across layers and formulates the per-layer scaling search as an optimisation problem. 
Zhou et al. \cite{Zhou2017inq} presented an incremental scheme that focuses solely on weights and pushes state-of-the-art networks below 6 bits with no loss of accuracy. However, all aforementioned works \cite{Gysel2016ristretto,Guo2017angeleye,Zhou2017inq} employ post-quantisation retraining 
to address losses in accuracy while pushing the designs to low-precision computations, assuming the availability of training data which is not always the case. 



FPGA-based CNN accelerators have been using 16-bit fixed-point precision for both activations and weights with minimal accuracy penalty \cite{Yufei_Ma_2017}\cite{Venieris_2017b} with Suda et al. \cite{suda2016throughput} employing 8-bit weights. Targeting applications where a significant error can be tolerated, optimised FPGA mappings of binarised \cite{Umuroglu_2017} and ternary networks \cite{Boucle2017ternaryfpga} have been proposed.

Following a different approach to the majority of existing quantisation methods, \textit{CascadeCNN} is relieved of the need to have access to the whole application dataset, as it does not employ 
model retraining in its quantisation scheme. Instead, a two-stage cascade structure is generated using only a small subset of labelled samples, where the higher-precision processing stage is responsible for restoring accuracy to user-defined acceptable levels. However, this cascade approach is orthogonal to methods which employ model fine-tuning and the proposed framework can be extended to accommodate a retraining step. In contrast to existing single-stage FPGA accelerators which rely on user-specified precision, \textit{CascadeCNN} introduces an automated quantisation method that takes into account the user-defined error tolerance and generates an optimised two-stage cascade architecture. Moreover, by utilising both DSPs and LUTs for its arithmetic units, \textit{CascadeCNN}'s FPGA-based architecture is designed to reach higher performance with every reduction in the wordlength.

 \begin{figure}
	\centering
	\includegraphics[trim ={105mm 62.5mm 29mm 64mm},clip, width=0.99\columnwidth]{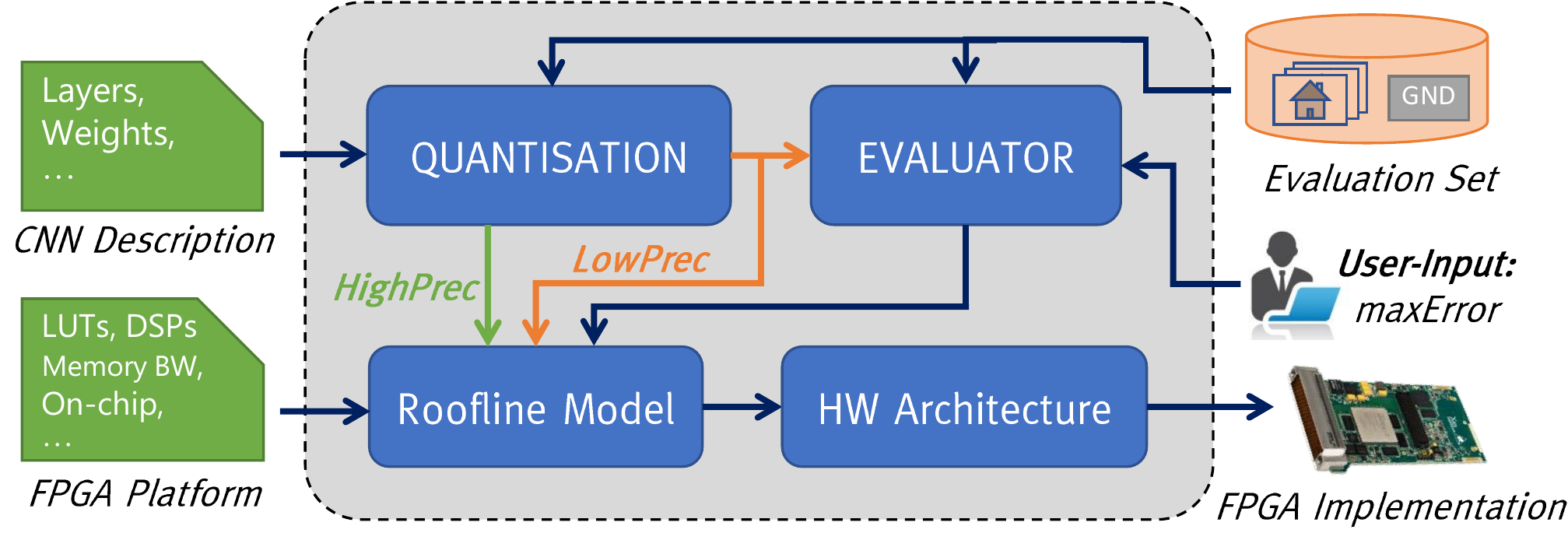}
	\vspace{-6mm}
	\caption{\textit{CascadeCNN}'s automated toolflow.}
		\vspace{-7mm}
	\label{fig:toolflow}
\end{figure}

\vspace{-1.1mm}
\section{CascadeCNN}
\vspace{-3.3mm}
\subsection{Overview}
\begin{figure}
 \vspace{-1mm}
	\centering
	\includegraphics[trim = {19mm 70mm 37.5mm 50.5mm},clip, width=0.90\columnwidth]{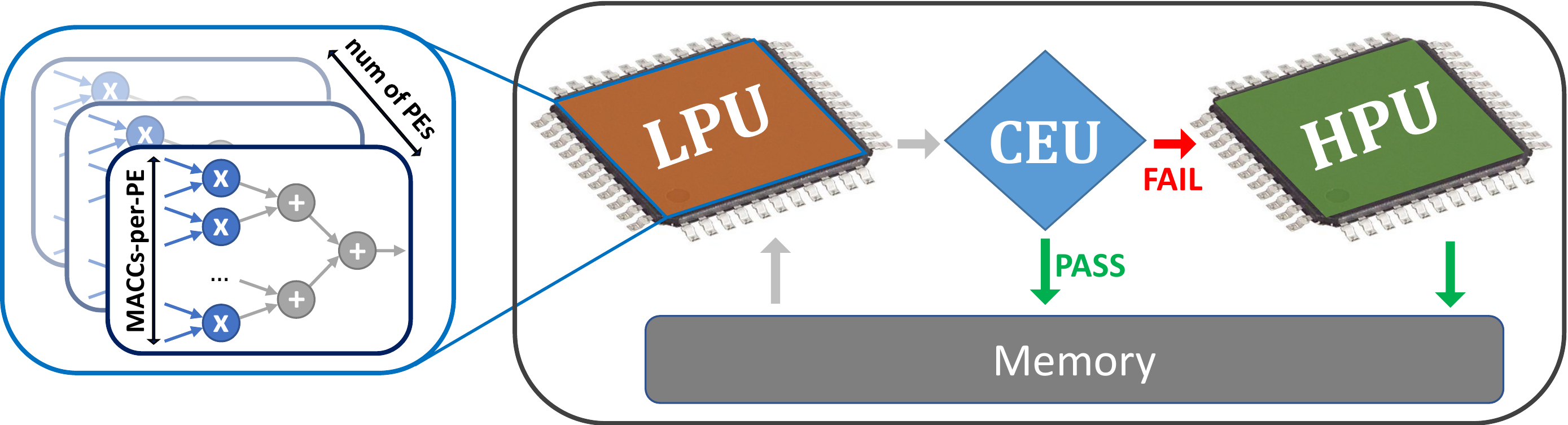}
	\vspace{-0.5mm}
	\caption{\textit{CascadeCNN}'s high-level architecture. }
	\label{fig:arch}
	\vspace{-6.5mm}
\end{figure}

\vspace{-1mm}
A high-level view of \textit{CascadeCNN}'s processing flow is depicted in Fig. \ref{fig:toolflow}. The framework is supplied with a high-level description of a trained CNN model (i.e. Caffe model), the available computational and memory resources of the target platform (LUTs, DSPs, Memory BW, on-chip and off-chip memory capacity) and an application-level error tolerance in a user-defined metric (e.g. top1/top5 classification error), along with a small evaluation set for the target application. 
\textit{CascadeCNN} searches the architectural design space and generates a two-stage hardware architecture, optimised for the particular CNN model and target device. The generated system for CNN inference (Fig. \ref{fig:arch}) consists of:

\begin{itemize}
 \item A low-precision unit (LPU) which employs low-precision arithmetic to trade lower accuracy with high throughput.
 \item A high-precision unit (HPU) which guarantees the same accuracy level as the reference model.
 \item A tunable Confidence Evaluation Unit (CEU) that detects samples that were wrongly classified by the LPU and redirects them to HPU for re-processing.
\end{itemize}

The key idea behind the proposed approach is that 
the LPU will process the whole workload, while the HPU will only process a fraction of it based on the CEU's evaluation of classification confidence on LPU's predictions, reducing its compute requirements. Moreover, the accuracy loss that is induced due to the extreme model quantisation of the LPU is restored to meet the user-specified error threshold.

\subsection{Quantisation}
\begin{figure}[h]
\vspace{-4.5mm}
	\centering
	\includegraphics[trim ={35mm 98mm 119mm 51mm}, clip, width=0.75\columnwidth]{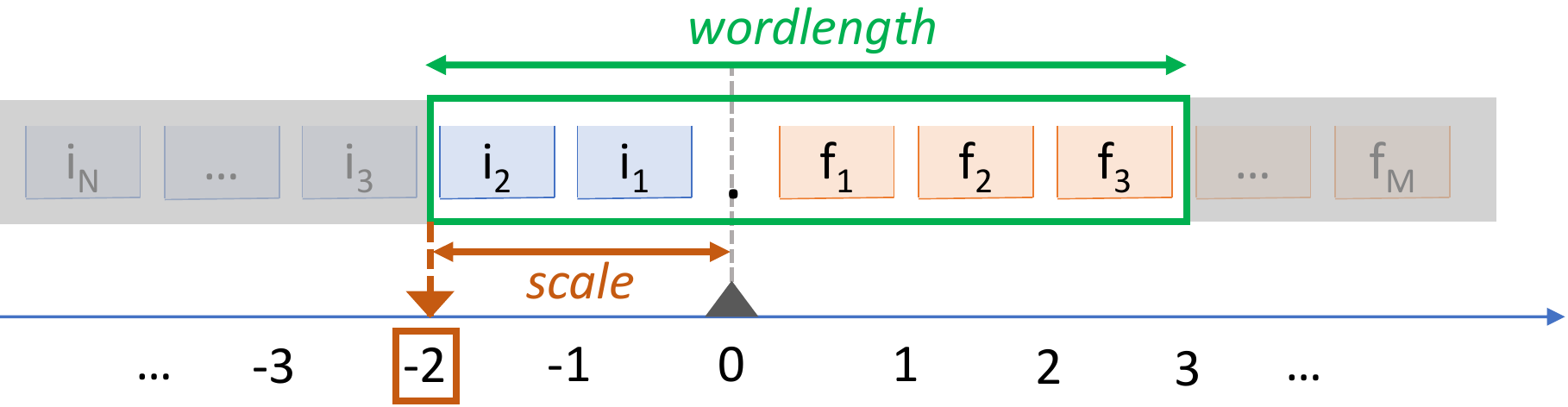}
	\vspace{-2mm}
	\caption{\textit{CascadeCNN}'s quantisation scheme parameters.}
	\label{fig:wordlength}
	\vspace{-3mm}
\end{figure}

\vspace{-1mm}
\label{sec:quant}
Arithmetic precision reduction is a widely studied technique which exploits the inherent redundancy of CNNs in order to reduce the memory bandwidth and footprint requirements, minimise power consumption and achieve higher performance. 
\textit{CascadeCNN} employs a fine-grained search space across possible precision quantisation schemes. This is formed by allowing different scaling factors for the weights and activations of each layer, which determine the ratio of integer and fractional bits, with a uniform wordlength across all layers (Fig. \ref{fig:wordlength}). 
The flexibility provided by this dynamic fixed-point approach 
enables smaller wordlengths to achieve high classification accuracy due to the different per-layer scaling factors, which would otherwise 
suffer significant accuracy loss if uniform scaling factors were employed across all layers. Moreover, employing a uniform wordlength for the whole network allows the derivation of a single hardware architecture that can be time-shared across all layers. The selected quantisation scheme of each layer can be described by the following tuple:
\vspace{-1.5mm}
\begin{equation}
    q_{layer} = <WL_{net}, SC^{weights}_{layer}, SC^{activations}_{layer}>
    \vspace{-1.5mm}
\end{equation}

\begin{figure}[t]
\vspace{-3.5mm}
\centering
\hspace{-1.3cm}
\subfloat[Uniform]{%
 	\includegraphics[trim={15mm 70mm 15mm 82mm},clip,width=0.55\columnwidth]{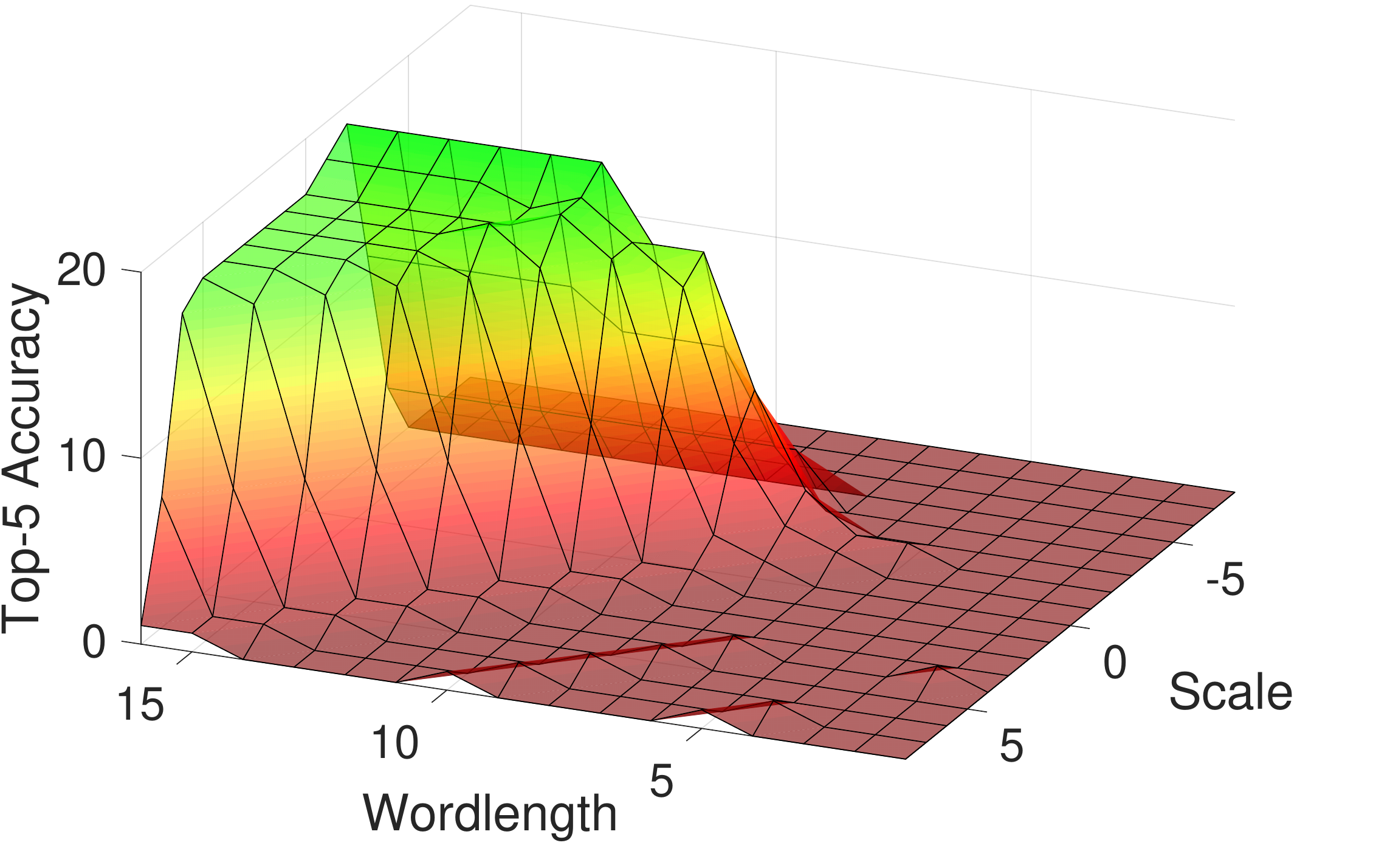}
  \label{fig:evaluation:revenue}%
}\qquad
\hspace{-1.2cm}
\subfloat[Single Layer]{%
 	\includegraphics[trim={0mm 75mm 0mm 80mm},clip,width=0.55\columnwidth]{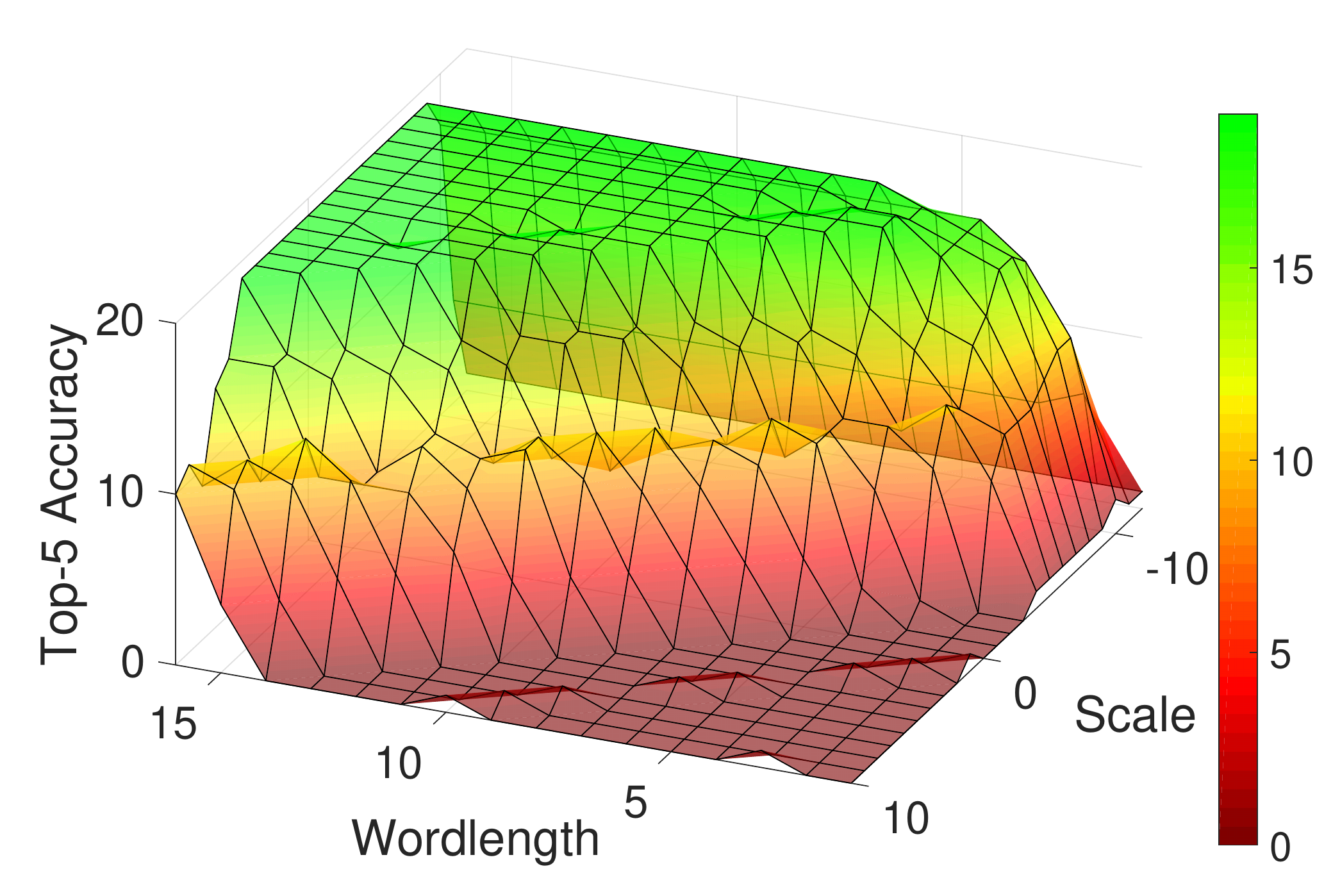}
  \label{fig:evaluation:avgPrice}%
}
\hspace{-10mm}
\vspace{-1mm}
\caption{Top-5 accuracy across the quantisation space, for \mbox{VGG-16} examined on 20 samples of ImageNet validation set.}
  \label{fig:quantSurf}%
  \vspace{-6mm}
\end{figure}

To determine the scaling factors for every explored wordlength of a given CNN, statistics regarding the quantisation impact of each layer on the application-level accuracy are extracted using a small user-provided evaluation set. Each layer's weights and activations are progressively quantised with decreasing wordlength values, examining a wide range of scaling factors, while computations and parameters for the rest of the network's layers employ floating-point representation. Fig. \ref{fig:quantSurf} illustrates snapshots of \textit{CascadeCNN}'s quantisation search space. The achieved classification accuracy of \mbox{VGG-16} is demonstrated across quantisation schemes on a dataset of 20 samples, for uniform scaling factors across the network (a), and for a single layer (b). As observed in this figure, useful conclusions about the dynamic range of each layer can be extracted even from a very small evaluation dataset. 

The extracted per-layer statistics are used to guide the exploration towards the combination of scaling factors that achieves the highest accuracy for each wordlength. In more detail, all quantisation schemes that exceed an accuracy threshold for each layer, are combined to form a search space for the particular network, which is heuristically searched to determine the classification scheme that achieves the highest network-level accuracy for the examined wordlength. In contrast to other frameworks, \textit{CascadeCNN} selects for the LPU an excessively-reduced wordlength quantisation scheme that achieves intermediate application-level accuracy, but with significantly higher performance when mapped on its custom precision-optimised hardware units. At run time, all input samples are processed by the LPU in order to obtain a rapid classification decision, which is then fed to the CEU. A wordlength that achieves an accuracy that complies with the user-specified error margins is selected for the HPU.
Provided that the reduced-precision model employed by the LPU of \textit{CascadeCNN} is derived from the given CNN by direct quantisation (without retraining), its run-time derivation at hardware level from the HPU's higher precision model is feasible. As a result of this weight-sharing approach, the model size and memory requirements of the proposed cascade approach remain the same as in the case of a single stage accelerator employing the precision selected for the HPU.

\vspace{-2mm}
\subsection{Confidence Evaluation Unit}
\begin{figure}
	\centering
	\includegraphics[trim =0mm 0mm 0mm 0mm, width=0.95\columnwidth]{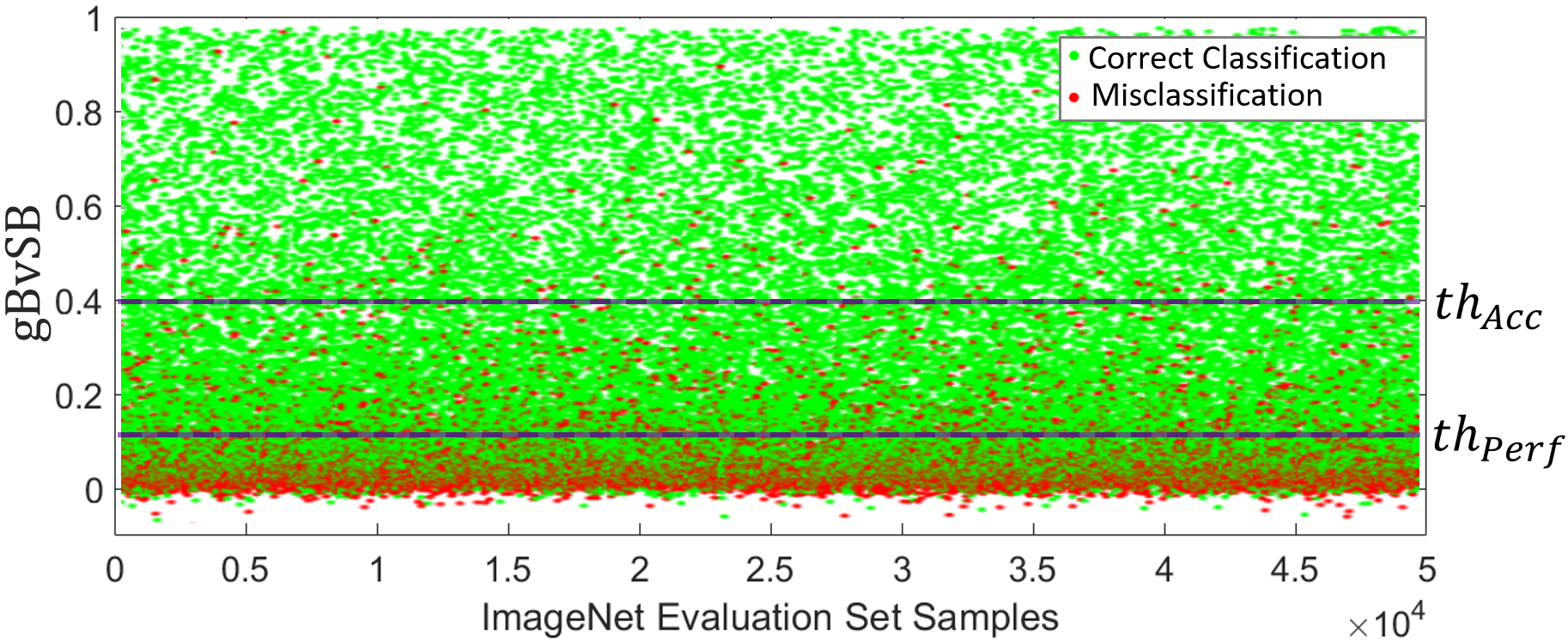}
	\vspace{-2mm}
	\caption{gBvSB$_{<5,10>}$ metric values across the ImageNet validation set for a VGG-16 5-bit LPU. 
	}
	\label{fig:bvsb}
	\vspace{-7mm}
\end{figure}

\vspace{-1mm}
\textit{CascadeCNN} allows the exploration of extreme quantisation schemes for the LPU, by aiming to identify potentially misclassified inputs based on the confidence of the LPU classification prediction. Low-confidence predictions, identified at run time by the CEU, are re-processed by the HPU to restore classification accuracy. To estimate the confidence of a prediction, we build on the work of \cite{joshi2009multi} by generalising the proposed Best-vs-Second-Best (BvSB) metric. BvSB was previously examining solely the distance between the two highest probability values, which is mostly applicable to binary classification, or when focusing exclusively on top-1 accuracy. Our generalised BvSB (gBvSB) metric is defined as:
\vspace{-2.5mm}
\begin{equation}
\vspace{-2.5mm}
  \text{gBvSB}_{<M,N>}(\boldsymbol{p}) = \sum_{i=1}^{M}p_i - \sum_{j=M+1}^{N}p_j
\end{equation}
where $p_i$ denotes the i-th element of the sorted probability vector $\boldsymbol{p} \in C$ of the predictions for $C$ classes and $M$ and $N$ are tunable parameters of gBvSB. An instance of this parameter space is illustrated in Fig. \ref{fig:bvsb}, indicating the gBvSB for VGG-16 predictions (quantised with the methodology described in Sec. \ref{sec:quant}) across the validation set of \mbox{ImageNet \cite{deng2009imagenet}}. In this context, a prediction is considered confident, and thus the processing ends at the low-precision unit, when:
\vspace{-2.3mm}
\begin{equation}
  \text{gBvSB}_{<M,N>}(\boldsymbol{p}) \ge th
  \vspace{-1.8mm}
\end{equation}
where $M$, $N$ and threshold $th$ form tunable parameters whose values are automatically determined by the proposed toolflow, using the provided evaluation set to meet the user-specified error tolerance. In this manner, the degree of uncertainty of the classification decision is based on how spiky the sorted probability distribution of a CNN's prediction is. As illustrated in Fig. \ref{fig:bvsb}, misclassified cases are uniformly distributed across the evaluation set. Hence, a small subset of these samples is adequate to provide an estimate of the percentage of misclassified cases that each CEU instance will fail to capture. In our experiments 200 samples\footnote{This value can be adjusted to meet required confidence intervals.} were used to tune the CEU's parameters to suit the user-specified error tolerance, out of the 1.2M samples consisting a typical ImageNet training set \cite{han2015deep}.  

It should be noted that a trade-off exists between the number of samples that were correctly classified by the LPU but are forwarded for re-processing to the HPU due to low confidence (false-negatives), and the number of misclassified samples that the CEU failed to identify due to the CNN's overconfident prediction (false-positives). The latter cases are responsible for the classification error that is introduced by \textit{CascadeCNN}, and can be reduced by setting the CEU parameters to a configuration that terminates the computation only on remarkably confident predictions ($th_{Acc}$). However, this would significantly increase the number of correctly classified inputs that are re-processed, which comes with a performance penalty on the system. In contrast, a configuration that only suggests re-processing on the extremely under-confident inputs, maximises performance at the cost of degraded classification accuracy ($th_{Perf}$). This trade-off (depicted in Fig. \ref{fig:bvsb}) is exploited to achieve the highest performance satisfying the user-specified error tolerance.

	\vspace{-1.5mm}
\subsection{Architecture}
	\vspace{-1.0mm}
\label{arch_sec}

A scalable, fine-grained hardware architecture is designed that is able to execute CNN inference, scale its performance with the resources of a target FPGA and exploit higher degrees of parallelism as the wordlengths of activations and weights decrease. The core of the architecture is a matrix-multiplication (\textit{MM}) unit, parametrised with respect to the tiling of each matrix dimension and the arithmetic precisions for activations and weights. The \textit{MM} unit comprises Multiply-Accumulate (MACC) units, grouped into Processing Elements (PEs) that perform dot-products (Fig. \ref{fig:arch}). By casting convolutions as matrix multiplications and using batch processing for Fully-Connected (FC) layers, both Convolutional (CONV) and FC layers are mapped on the \textit{MM} unit. 
Particularly, every CONV layer can be described as:
	\vspace{-2.5mm}
\begin{equation}
CONV< H,W,N_{IN}, N_{OUT}, K_H, K_W, S_H, S_W, Z>
	\vspace{-2.5mm}
\end{equation}
where $H$ and $W$ denote the height and width of each input feature map, $N_{IN}$ and $N_{OUT}$ the number of input and output feature maps respectively, $K_H$ and $K_W$ the height and width of the layer's kernels, $S_H$ and $S_W$ the stride size in the height and width directions of each input respectively and $Z$ the zero-padding on the input feature maps. FC layers can be described similarly to CONV layers under the following formulation:
	\vspace{-2.0mm}
\begin{equation}
FC = CONV< 1,1,N_{IN}, N_{OUT}, 1, 1, 1,1, 0>
	\vspace{-2.0mm}
\end{equation}

\begin{figure}[t]
	\centering
	\includegraphics[trim =4mm 10mm 4mm 33mm,clip, width=\columnwidth]{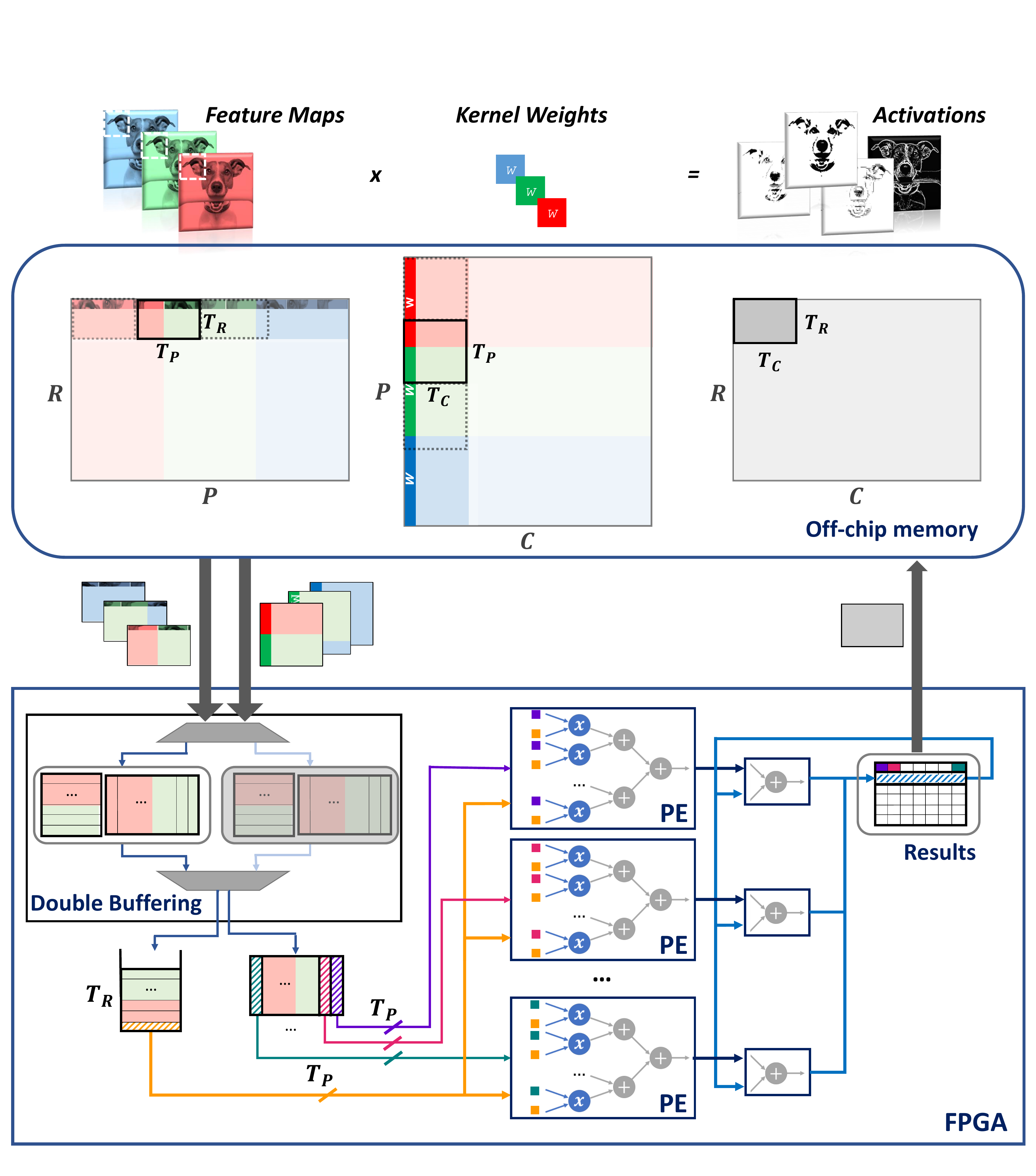}
		\vspace{-6mm}
	\caption{Overview of proposed hardware architecture.}
	\label{fig:hw_arch}
	\vspace{-6mm}
\end{figure}

For each CONV layer, a $R_{CONV}$$\times$$P$ input activations matrix is constructed to represent its input feature maps. The input activations matrix is organised with each row containing the unrolled feature map values for a single sliding window position, concatenated for all input channels. The number of sliding window positions of a layer can be calculated as:
	\vspace{-1.5mm}
\begin{equation}
\resizebox{0.9\linewidth}{!}{
 $R_{CONV} = \ceil*{\frac{H +2Z-(K_H-1)}{S_H}} \ceil*{\frac{W +2Z-(K_W-1)}{S_W}}$
 }
 	\vspace{-1.5mm}
\end{equation}
while the number of elements included in the unrolled sliding window for all channels of the input feature map volume is $\mathbf{P = K_H K_W N_{IN}} \inlineeqnum\label{eq:Pval}$. In the case of FC Layers, batching is employed to form a similar $R_{FC}$$\times$$P$ matrix, each row of which contains the input feature vector of size $P$ for a different input sample, and hence: $\mathbf{ R_{FC} = BatchSize \inlineeqnum\label{eq:RFC}}$. Similarly each layer's weights are unrolled to form a $P$$\times$$C$ weights matrix, with each of the $C$ columns corresponding to a different kernel, expressed as a $P$-element vector of concatenated unrolled values for $N_{IN}$ channels. Hence, the number of columns of $P$$\times$$C$ matrix is identical to the number of output feature maps of the layer: $\mathbf{ C = N_{OUT} \inlineeqnum}$. As a result of the multiplication between the $R$$\times$$P$ and $P$$\times$$C$ matrices, a $R$$\times$$C$ matrix is produced, with each column representing the output feature map values produced by the different filters of the layer (Fig. \ref{fig:hw_arch}-top). 


The \textit{MM} unit employs tiling across all three dimensions ($R,P,C$) with tile sizes of $T_R$, $T_P$, $T_C$ accordingly. Input tiles $T_R$$\times$$T_P$ and $T_P$$\times$$T_C$ are loaded from the off-chip memory employing double buffering to hide the memory latency between the processing of different tiles. Moreover, to sustain a high memory bandwidth utilisation in the case of quantised weights and activations, the low-precision values are packed into larger chunks that match the memory ports width (e.g. 4-bit values packed in chunks of 16 for 64-bit ports).

Fig. \ref{fig:hw_arch} shows the proposed parametrised hardware architecture. A tile of the output activations matrix ($T_R$$\times$$T_C$) is computed by accumulating the results of $\ceil*{\frac{P}{T_P}}$ tile multiplications. 
Intermediate results are kept on-chip, and only the final output tiles are transferred back to the off-chip memory. Each tile multiplication consists of a set of dot-product operations. The proposed architecture exploits two different levels of parallelism to allow for a fine-grained architectural search space. To exploit the fine-grained parallelism in dot-products, the dot-product between $T_P$-element vectors is fully unrolled and mapped on a PE that contains a multiplier array followed by an adder tree. At the same time, $T_C$ PEs are instantiated concurrently, to parallelise the dot-products between all columns of the $T_P$$\times$$T_C$ tile with a single row of the $T_R$$\times$$T_P$ tile. Finally, the rows of the $T_R$$\times$$T_P$ tile are processed in a pipelined manner in order to maximise throughput. 

\noindent
\textbf{Precision-aware mapping of PEs to FPGA resources.}
In contrast to the majority of the existing FPGA-based designs, the proposed architecture utilises both LUTs and DSPs to implement its MACC units. A similar strategy was partially explored in \cite{Yufei_Ma_2017}, focusing solely on 16-bit operands. With smaller wordlengths being less LUT-costly, employing LUT-based arithmetic units alongside DSP-based units enables the proposed architecture to reach higher performance by instantiating higher number of MACC units as the wordlength decreases. Moreover, for the LPU, we introduce a strategy that packs a pair of low-precision MACCs (less than or equal to 5 bits) on a single $25$$\times$$18$ DSP, by positioning targeted zero guard-bits between the input operands to avoid interference on the results and, in this way, doubling the computational capacity of the FPGA's DSP blocks. This technique is enabled by the extreme quantisation employed by the LPU, 
since larger wordlengths would either limit its applicability only to MACC operations with a shared operand \cite{Nguyen2017DoubleMD}, or restrict it completely.

\vspace{-2.6mm}
\subsection{Design Space Exploration}
\vspace{-1mm}
\label{sec:dse}

Given a CNN-FPGA pair and a particular wordlength, \textit{CascadeCNN} searches the architectural design space by means of a roofline-based performance model \cite{williams2009roofline} in order to determine the highest performing configuration of the architecture. The configurable parameters comprise the matrix tile sizes, that determine the number of PEs and MACCs-per-PE as described in Sec. \ref{arch_sec}, along with a tiled batch size, to be discussed in Sec. \ref{batch_size_sec}. In this manner, \textit{CascadeCNN} generates two architectures, the LPU and the HPU, which are optimised for different wordlengths for the provided CNN-FPGA pair.

\subsubsection{Tile Sizes}
Having parametrised the \textit{MM} operation and its underlying architecture accordingly, the design space is explored to determine the highest performing tile size tuple $<$$T_R, T_P,T_C$$>$ for each CNN-FPGA-precision instance. A per-layer analysis is initially performed, the results of which are combined to yield a shared architecture across all layers, which maximises the overall throughput. To enable fast and exhaustive DSE, a roofline model is developed that associates the predicted performance of each architectural configuration with its operational intensity (which provides an estimate of the ratio between computational workload and memory traffic), to obtain the highest performing design point.

  
The attainable performance of the architecture given a triplet of tile sizes  $<T_R, T_P,T_C>$ is calculated as:
\vspace{-1.4mm}
\begin{equation}
\resizebox{0.89\linewidth}{!}{
    $Perf(ops/s) = \frac{workload}{II} = \frac{2 R P C }{\ceil*{\frac{R}{T_R}} \ceil*{\frac{P}{T_P}}  \ceil*{\frac{C}{T_C}} T_R } clk(WL)$
    }
\vspace{-1.4mm}
\end{equation}
where $workload$ denotes the total number of operations for a single input of the particular layer in ops/input, with each MACC counting for two operations, while the Initialisation Interval $II$ denotes the number of cycles required to process an input. Finally, $clk(WL)$ is the achieved clock frequency of the architecture, depending on the selected wordlength.

As can be seen in Algorithm \ref{alg:mm_mult} (describing the tiled \textit{MM} approach) the rate of read and write memory transactions varies throughout the computation. To deal with this imbalance, the performance analysis is configured to treat each iteration of \textit{loop2} as a single step, capturing the total memory traffic and processing load required per 
output tile ($T_R$$\times$$T_C$). In order to serve the fine-grained quantisation space, the roofline model is adjusted to employ bits, instead of bytes, as its data measurement unit. Hence, the operational intensity of the architecture for each such step is formulated as:
\vspace{-1.5mm}
\noindent
\begin{equation}
\resizebox{0.89\linewidth}{!}{
    $opInt(ops/bit) = \frac{ops}{memAcc} = \frac{2 T_R P T_C }{(T_R P + P T_C + T_R T_C)WL}$
    }
    \vspace{-1.5mm}
\end{equation}
where $ops$ denotes the total amount of operations required for the computation of an output tile ($ops/tile$ in a single iteration of \textit{loop2}), while $memAcc$ denotes the number of bits that needs to be transferred between the off-chip memory and the FPGA in bits/tile, during the same computation. 
  
By enumerating the possible tile size combinations and calculating the corresponding $Perf$ and $opInt$ values, the design space is formed for the examined CNN layer and precision. Considering the target FPGA resources, only a subspace of this design space corresponds to platform-supported architectures. The limiting factors of the platform are its peak attainable performance, obtained by resource modelling for a wide range of precisions, and memory bandwidth. Specifically, MACCs are implemented on both DSPs and LUTs (Sec. \ref{arch_sec}), hence:
\vspace{-2.95mm}
\begin{equation}
\resizebox{0.89\linewidth}{!}{
    $totalOps = 2 \Big( \frac{availLUT}{LUT_{MACC}(WL)} + availDSP \cdot MACC_{DSP}(WL) \Big) $
    }
    \vspace{-0.6mm}
\end{equation}
where $availLUT$ and $availDSP$ denote the number of available resources on the target device, while $LUT_{MACC}(WL)$ indicates the precision-specific resource consumption of a single MACC unit when implemented on LUTs and $MACC_{DSP}(WL)$ indicates whether packing of multiple MACCs on a single DSP is supported by the particular wordlength. With respect to the on-chip storage, the on-chip memory requirements for each design point is obtained as:
\vspace{-1.6mm}
\begin{equation}
\vspace{-1.6mm}
\resizebox{0.85\linewidth}{!}{
    $onChipMem(bits) = 2 (T_R T_P + T_P T_C + T_R T_C) WL$
    }
\end{equation}
where the term 2 accounts for double buffering.

Design points exceeding the platform's computational roof or on-chip memory capacity are excluded from the platform-supported design space. Additionally, for design points whose bandwidth requirement exceeds the available, their attainable performance is projected to meet the platform's memory roof. 

  \begin{center}
  \vspace{-3mm}
    \scalebox{0.75}{
    \begin{minipage}{1\linewidth}

\begin{algorithm}[H]

\caption{CNN Layer as Tiled Matrix Multiplication}
\label{alg:mm_mult}
\begin{algorithmic} 
\FOR[(loop1)] { $r=1$ \textbf{to} $\ceil*{\frac{R}{T_R}}$ \textbf{step} $T_R$} 
\FOR[(loop2)] { $c=1$ \textbf{to} $\ceil*{\frac{C}{T_C}}$ \textbf{step} $T_C$} 
\STATE $RegRC \leftarrow zeros(T_R$$\times$$T_C)$
\FOR[(loop3)] { $p=1$ \textbf{to} $\ceil*{\frac{P}{T_P}}$ \textbf{step} $T_P$}
\STATE $RegRP \leftarrow memRead(T_R$$\times$$T_P)$
\STATE $RegPC \leftarrow memRead(T_P$$\times$$T_C)$
\FOR[(loop4)] { $rr=1$ \textbf{to} $T_R$ \textbf{step} $1$}
\FOR[(loop5)] { $cc=1$ \textbf{to} $T_C$ \textbf{step} $1$}

\STATE $RegRC(rr,cc) \textbf{+=}$ \\
\begin{flushright}
  $ dot\_product(RegRP(rr),RegPC(cc)) $
\end{flushright}
\ENDFOR
\ENDFOR
\ENDFOR
\STATE $memWrite(RegRC)$
\ENDFOR
\ENDFOR

\end{algorithmic}

\end{algorithm}

 \end{minipage}
}
\end{center}  

After completing the DSE for every layer of a given CNN using a particular precision, a single tile size triplet that maximises the overall performance across all layers needs to be determined. Nevertheless, since significant variability is observed in the matrix dimensions of different layers, different tile sizes yield the highest performance for each layer. 
This can be explained by the fact that deeper layers tend to have more and smaller filters in contrast with the first layers that consist of fewer filters, operating on larger feature maps. To calculate the average attainable performance of a design point with fixed tile sizes across all layers, a weighted average based on the percentage of processing cycles each layer is contributing for the computation of a single batch of inputs is employed:
\vspace{-2mm}
\begin{equation}
\vspace{-2mm}
\resizebox{0.82\linewidth}{!}{
    $ovrlPerf(ops/s) = \frac{\sum_{i=1}^{N_{layers}}{k(i) Perf(i) II(i)}}{\sum_{i=1}^{N_{layers}}{II(i)}}$,
    }
\end{equation}
\begin{equation}
\vspace{-2 mm}
\resizebox{0.62\linewidth}{!}{
    $k(i) = 
    \begin{cases} 
        BatchSize  \text{ , if  $Layer_i$ is CONV} \\ 
        1  \  \ \ \ \ \ \ \ \ \ \ \   \text{       , if  $Layer_i$ is FC}
    \end{cases}$
    }
\end{equation}
where $II(i)$ is the number of cycles for processing a single input sample by the $i$-th layer. The platform-supported design point achieving the maximum $ovrlPerf$ is selected for the implementation of the examined CNN-FPGA-precision triplet. 

\subsubsection{Batch Size}
\label{batch_size_sec}
As described in Sec. \ref{arch_sec}, input samples are processed by \textit{CascadeCNN} in batches, given that this work focuses on throughput optimisation. All samples of a batch are first processed by the LPU, with the CEU estimating the confidence of each prediction. Subsequently, the FPGA is reconfigured with the optimised hardware architecture for the precision employed by the HPU, which processes a fraction of the input samples, based on the CEU's response. The reconfigurability of FPGAs enables highly optimised -for the selected LPU and HPU wordlengths- processing units to alternate with each other, while occupying nearly all the available resources of the target platform to exploit more parallelism. Nevertheless, each full reconfiguration of the device introduces substantial delay that can aggravate the overall system performance. Employing large batch sizes, so that multiple samples are processed by each processing unit between each reconfiguration, compensates for the cost of reconfiguring the device (as reconfiguration time becomes negligible compared to the overall processing time of the batch). Thus, the largest $BatchSize$ that can be accommodated by the off-chip memory (considering both inputs and intermediate results) is selected.

 
However, large $BatchSize$ values affect the $R$ dimension of FC layers' input activations matrices, which can aggravate the load imbalance between layers, and thus degrade the average performance of the derived architecture. To address this issue, tiling of the batch size is employed. A corresponding tile size $T_{Batch}$ is further parametrising the design space to ensure better load balance across layers, with $T_{Batch}$ replacing $BatchSize$ in Eq. (\ref{eq:RFC}). To preserve the capability of exhaustively searching the design space, only values that are 
powers of two or multiples of 1024 are examined for $T_{Batch}$. The tiling factor of the system's batch size is selected to maximise the average attainable performance. 
The resulting system's LPU processes the whole $BatchSize$ in groups of $T_{Batch}$ samples, before switching to the HPU, in order to achieve better load balance between CONV and FC layers.
 
\vspace{-1.6mm}
\section{Evaluation}
\vspace{-1.5mm}
\subsection{Experimental Setup}
\vspace{-1.5mm}
In our experiments, we target image classification as our application case study, using pretrained models on the \mbox{ImageNet \cite{deng2009imagenet}} dataset. \textit{CascadeCNN} is provided with models of AlexNet \cite{krizhevsky2012imagenet} and VGG-16 \cite{simonyan2014very}, along with a small subset of the ImageNet validation set as an evaluation set (200 labelled samples, out of 50,000). Experiments are conducted on a wide range of values for the user-specified parameter of error tolerance. Matlab 2017a is used to investigate the quantised fixed-point network behaviour and determine the highest achieving quantisation scheme for each wordlength as described in Sec. \ref{sec:quant}, as well as to obtain network predictions on the evaluation set to tune the CEU parameters. All hardware designs were synthesised and placed-and-routed using the Xilinx Vivado HLS and Vivado Design Suite (v17.2) and evaluated on the Xilinx Zynq ZC706 and UltraScale+ ZCU102 boards. The ARM CPU was used to execute the softmax layer at the output of each CNN and the CEU. 
\vspace{-1.6mm}
\subsection{Architectural Performance and Accuracy}
\begin{figure}
  	\vspace{-2.8mm}
	\centering
	\includegraphics[trim ={108mm 70mm 108mm 72mm}, clip, width=0.90\columnwidth]{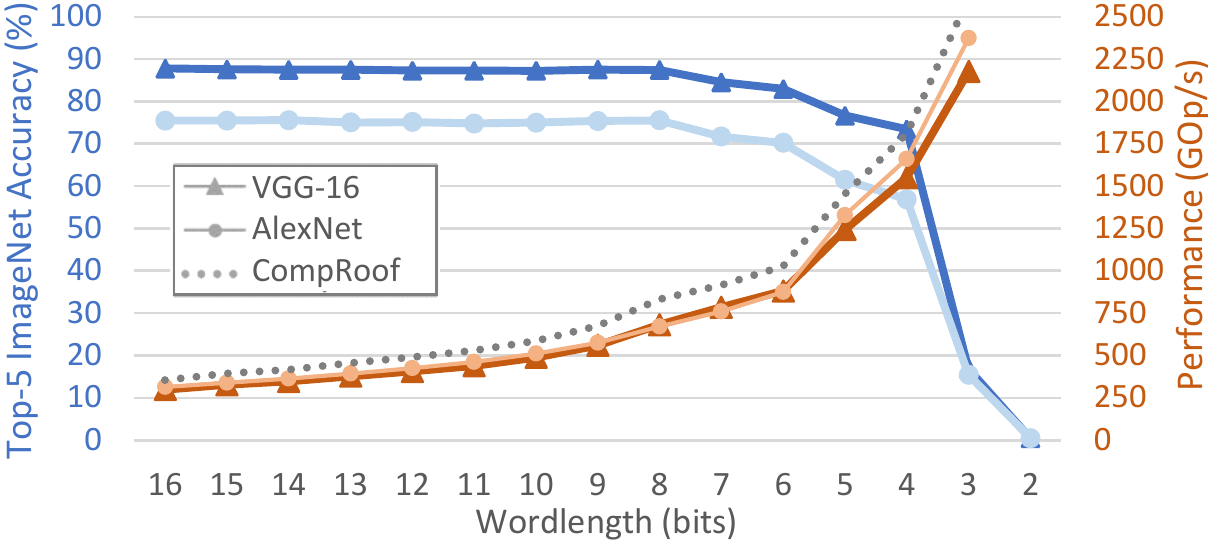} 
		    \vspace{-1.8mm}
	\caption{ Achieved classification accuracy and measured performance as a function of wordlength on Zynq ZC706. }
	\label{fig:prec}
		    \vspace{-7mm}
\end{figure}

\vspace{-1mm}
In this section, the performance and accuracy of the proposed architecture implementing the \textit{CascadeCNN}'s processing units, is evaluated as a function of employed wordlength. For both AlexNet and VGG-16, \textit{CascadeCNN} yields a wordlength of 4 bits for the LPU, by selecting the smallest wordlength that did not experience catastrophic accuracy loss during the quantisation scheme exploration.  
This 4-bit LPU introduces a 14.38\% and 18.65\% degradation in classification accuracy compared to a 16-bit precision respectively (Fig. \ref{fig:prec}). It should be noted that (similarly to what has been reported in \cite{suda2016throughput}) negligible accuracy variability is observed in the range between 16-bit and 8-bit wordlength implementations, provided by the employed quantisation methodology. As the resulting architectures for higher-precision units in the range of 16 bits demonstrate degraded performance due to increased $LUT_{MACC}$ requirements and lower achieved clock frequencies, an 8-bit implementation is selected to act as a baseline for \textit{CascadeCNN}'s error. The CEU parameters are tuned on the evaluation dataset to generate systems that introduce a wide range of classification errors, compared to the 8-bit baseline. 

Using the roofline performance analysis of \mbox{Sec. \ref{sec:dse}}, the design points achieving the highest performance 
for AlexNet and VGG-16, for a wide range of wordlengths are extracted. The accuracy of the developed roofline model is evaluated by comparing the predicted to the real, measured performance for various points across the design space, demonstrating an average error of 6.8\% (geo. mean). 
Fig. \ref{fig:prec} shows the measured performance on the ZC706 board and the achieved accuracy across multiple wordlengths. The fine granularity of the architectural design space allows the selected design points for all examined precisions to approach the device's computational roof. The 4-bit LPU architectures achieve a throughput improvement of 2.28$\times$ for VGG-16 and 2.48$\times$ for AlexNet compared to the faithful, zero-error 8-bit architecture\footnote{
AlexNet, 4-bit: [DSPs:100\%, LUTs:80.6\%, BRAM:5.16\%]\\
\phantom{000}AlexNet, 8-bit: [DSPs:100\%, LUTs:83.1\%, BRAM:4.34\%]\\
\phantom{000}VGG-16, 4-bit: [DSPs:100\%, LUTs:81.4\%, BRAM:4.88\%]\\
\phantom{000}VGG-16, 8-bit: [DSPs:100\%, LUTs:82.9\%, BRAM:5.76\%]
} and a speed-up of 5.18$\times$ for VGG-16 and 5.29$\times$ for AlexNet compared to the 16-bit counterpart, which is the most widely used precision by existing FPGA \mbox{accelerators \cite{Yufei_Ma_2017}\cite{Venieris_2017b}.}

\vspace{-2.0mm}
\subsection{End-to-End Cascade Performance}
\vspace{-1.5mm}
To evaluate the performance gains of \textit{CascadeCNN}, we compare the generated two-stage system for different error tolerance values with a baseline single-stage architecture. \textit{CascadeCNN}'s fine granularity of CEU search space enables it to fully exploit the performance-accuracy trade-off supporting arbitrary classification accuracy, in contrast with the coarser precision-accuracy trade-off that forms the only tunable parameter of the straight-forward quantisation approach adopted by the baseline. To address this incompatibility of design points across the accuracy dimension, each \textit{CascadeCNN} instance is compared with a baseline HPU, optimised for a precision that achieves the same or better accuracy as the cascade system (ranging from 5 to 8-bit wordlengths). \textit{CascadeCNN}'s HPU adopts the same precision as the baseline, whereas its LPU maintains the 4-bit wordlength yielded by the toolflow. For the cascade architecture, the overall measured processing time for a batch of inputs includes LPU's and CEU's latency for the whole batch, HPU's latency for the re-classified samples and the FPGA's reconfiguration time. 

\begin{figure}
  	\vspace{-3.7mm}
	\centering
	\includegraphics[trim ={73mm 70mm 73mm 49mm},clip, width=\columnwidth]{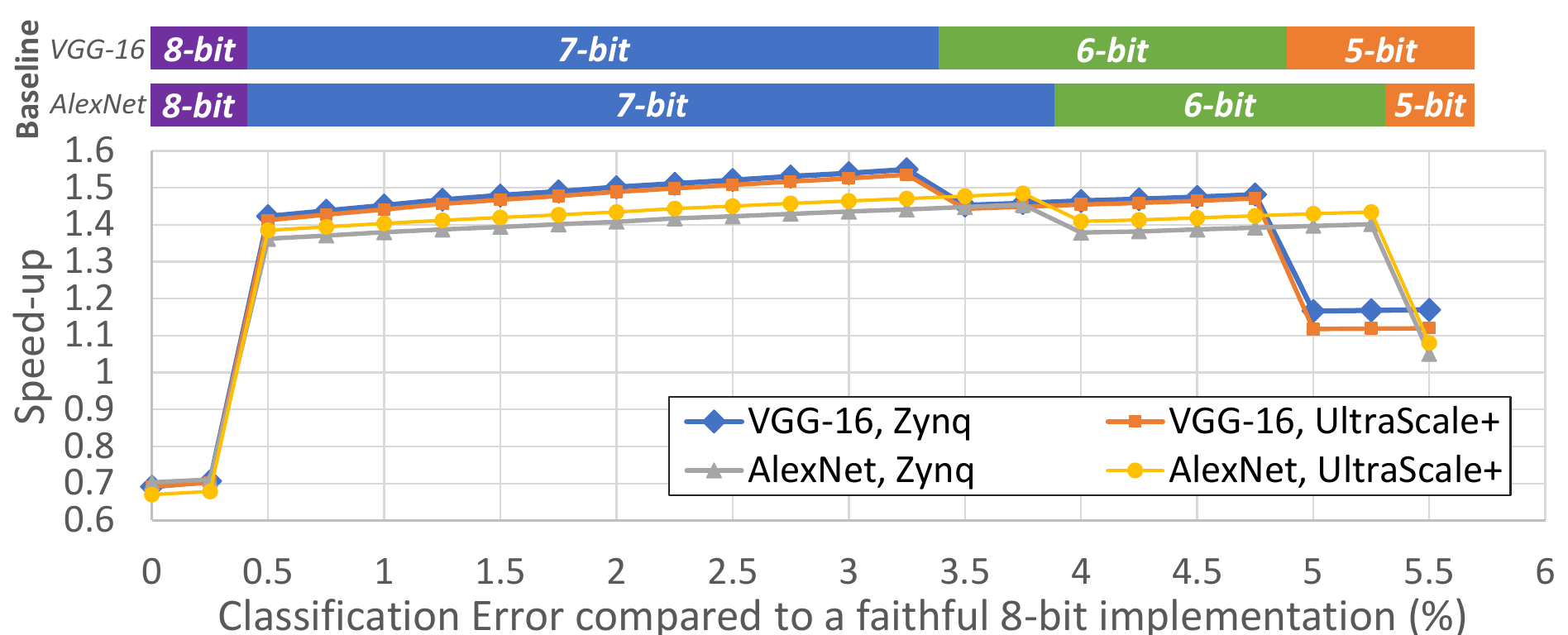}
			    \vspace{-5.5mm}
	\caption{\textit{CascadeCNN} Speed-up as a function of error tolerance.}
	\label{fig:speedup}
			    \vspace{-6.8mm}
\end{figure}

The achieved throughput gain is depicted in Fig. \ref{fig:speedup} across a wide range of error thresholds on ZC706 and ZCU102 boards. When zero or extremely small error tolerance (below 0.25\%) is required by the user, the CEU adopts a strict evaluation policy that results in an excessively high percentage of rejected samples forwarded to the HPU for re-processing. This introduces significant computational load that results in a slow-down of the overall cascade architecture. In such cases, where the baseline architecture outperforms the two-stage design, \textit{CascadeCNN} generates an optimised single-stage architecture, that meets the required error tolerance. For intermediate error levels (0.5\% - 5\%) in both target platforms, the proposed cascade system outperforms the baseline by up to 48\% for AlexNet and 55\% for VGG-16, for the same resource budget and accuracy. Finally, in the case of high error tolerance (over 5\%), the speed-up becomes less significant as the gap in wordlength between the LPU and the baseline design closes.

A persistent gain on throughput is observed across the two target devices, which comes with the scalable performance obtained by the highly parametrisable architecture. Moreover, although resource-rich devices, such as ZCU102, are burdened by larger reconfiguration time, \textit{CascadeCNN}'s DSE alleviates that cost by 
employing larger batch sizes, that allow more rare reconfigurations and hence higher amortisation of their cost. 

The proposed toolflow can also employ other existing CNN accelerator architectures as a basis for its LPU and HPU components, with variable performance gains. The CEU component is executed together with the softmax layer on the CPU, with their aggregate processing time being 4$\times$ faster than the FPGA-side computations. With the FPGA and the CPU acting as a processing pipeline across the batches of inputs, the latency of the CPU side is hidden and the system's overall performance is determined by the FPGA computations. 

\vspace{-2.5mm}
\subsection{Comparison with Existing FPGA Work}
\vspace{-1.5mm}
This section explores the performance of the proposed framework with respect to existing FPGA work that does not consider model retraining. This is investigated by comparing with a set of state-of-the-art works, including the highest performing AlexNet and VGG-16 on Zynq 7045 \cite{Venieris_2017b}\cite{Guo2017angeleye}, on Arria 10  \cite{Aydonat_2017}\cite{Zhang2017fpga}, the Escher accelerator that optimises memory bandwidth utilisation \cite{Yongming2017} and an RTL-based automated CNN-to-FPGA toolflow \cite{Yufei_Ma_2017}. Table \ref{table:otherworks} presents the measured performance results. The performance column lists the throughput in GOp/s and the performance density in GOp/s/DSP (shown in brackets). For each network, three \textit{CascadeCNN} design points are examined (C1,C3,C5), generated by the toolflow by setting the error tolerance to 1\%, 3\% and 5\%.

\textbf{HPU Performance.} To evaluate the architecture of the HPU, each accelerator is compared with the HPU of the same wordlength. In this respect, the 16-bit AlexNet HPU outperforms the existing designs of \cite{Venieris_2017b} and \cite{Yongming2017} in both achieved GOp/s and GOp/s/DSP and reaches 78\% of the GOp/s/DSP of the highly-optimised, hand-tuned design of \cite{Aydonat_2017}. With respect to VGG-16, the 16-bit HPU outperforms \cite{Venieris_2017b}, \cite{Guo2017angeleye} and \cite{Yufei_Ma_2017} and reaches 56\% of the performance density of the mixed OpenCL-RTL design of \cite{Zhang2017fpga}. 

\textbf{CascadeCNN Performance.} The two-stage cascade architecture however, achieves up to 3.31$\times$ higher performance density on AlexNet compared to the hand-tuned architecture of \cite{Aydonat_2017}, with an average of 3.1$\times$ (geo. mean) in the error tolerance range of 1 to 5\%. In the same error range, \textit{CascacdeCNN} also outperforms the works of \cite{Venieris_2017b} and \cite{Yongming2017} by up to 8.29$\times$ and 24.83$\times$ respectively. 
Similarly for VGG-16, the full cascade architecture outperforms the mixed OpenCL-RTL design of \cite{Zhang2017fpga} by up to 2.73$\times$ with an average of 2.36$\times$ (geo. mean) in the error tolerance range 1 to 5\%, while demonstrating remarkable gains compared to all other examined designs. Overall, in cases that tolerate an extra error of 1\% to 5\%, \textit{CascadeCNN} demonstrates significant speed-ups with respect to the state-of-the-art existing accelerators.

\begin{table}
\centering
\caption{Comparison with Existing FPGA Work}
\vspace{-2mm}
\resizebox{\columnwidth}{!}{
\setlength\tabcolsep{5pt} 

\begin{tabular}{ cc|ccc|cc} 
     \hlineB{2} 
     \textbf{Work} &\textbf{FPGA}  & \textbf{WL}  & \textbf{Freq} & \textbf{Performance$^1$} &  \textbf{Speed-up$^2$}  & \textbf{Err} \\
     \hlineB{2} 
	\multicolumn{2}{c}{\textbf{\phantom{0}AlexNet} } &\multicolumn{5}{c}{\phantom{0} }  \\
	\hlineB{2} 
	HPU & Z-7045 & 16-bit & 131 MHz & \phantom{0}313.48\phantom{0}(0.35) & \textbf{1.00} & 0\% \\ 
	\cite{Venieris_2017b} & Z-7045& 16-bit & 125 MHz & \phantom{0}161.98\phantom{0}(0.18) & 0.51 & 0\% \\
	\cite{Yongming2017}  & Virtex-7 & 16-bit & 100 MHz & \phantom{0}135.00\phantom{0}(0.06) & 0.17 & 0\% \\
	\cite{Aydonat_2017}  & Arria10$^\ddagger$ & 16-bit & 303 MHz & 1382.00\phantom{0}(0.45) & 1.29 & 0\% \\
	\hline
	C1$^\star$  & Z-7045 & [4,7]bit$^\dagger$  & 150 MHz & 1190.35\phantom{0}(1.32) & 3.77 & 1\% \\ 
	C3$^\star$   & Z-7045 & [4,7]bit$^\dagger$  & 150 MHz & 1238.66\phantom{0}(1.37) & 3.91 & 3\% \\ 
	C5$^\star$   & Z-7045 & [4,6]bit$^\dagger$  & 150 MHz & 1343.61\phantom{0}(1.49) & 4.26 & 5\% \\
	\hlineB{2} 
	\multicolumn{2}{c}{\textbf{\phantom{0}VGG-16} } &\multicolumn{5}{c}{\phantom{0} }  \\    
	\hlineB{2} 
	HPU  & Z-7045& 16-bit & 131 MHz & \phantom{0}299.07\phantom{0}(0.33) & \textbf{1.00} & 0\% \\
	\cite{Venieris_2017b} & Z-7045 & 16-bit & 125 MHz & \phantom{0}123.12\phantom{0}(0.14) & 0.42 & 0\% \\
	\cite{Guo2017angeleye}  & Z-7045 & 16-bit & 150 MHz & \phantom{0}136.97\phantom{0}(0.15) & 0.45 & 0\% \\
	\cite{Yufei_Ma_2017}  & Arria10$^\ddagger$ & 16-bit & 200 MHz & \phantom{0}619.13\phantom{0}(0.23) & 0.70 & 0\% \\
	\cite{Zhang2017fpga}  & Arria10$^\ddagger$ & 16-bit & 385 MHz & 1790.00\phantom{0}(0.59) & 1.79 & 0\% \\
	\hline
	HPU  & Z-7045 & 8-bit & 150 MHz & \phantom{0}680.91\phantom{0}(0.75) & 2.27  & 0\% \\
	\cite{Guo2017angeleye} & Z-7045 & 8-bit & 150 MHz & \phantom{0}273.76\phantom{0}(0.30)* & 0.91 & 0\%\\
	\hline
	C1$^\star$    & Z-7045 & [4,7]bit$^\dagger$  & 150 MHz & 1140.28\phantom{0}(1.26) & 3.82 & 1\% \\ 
	C3$^\star$    & Z-7045 & [4,7]bit$^\dagger$  & 150 MHz & 1208.69\phantom{0}(1.34) & 4.06 & 3\% \\ 
	C5$^\star$    & Z-7045 & [4,6]bit$^\dagger$  & 150 MHz & 1450.13\phantom{0}(1.61) & 4.88 & 5\% \\
	\hlineB{2} 
	\multicolumn{7}{c}{ * projected,\phantom{2}  $\dagger$ denotes [LPU,HPU] precision pair,\phantom{2}   $\ddagger$ refers to GX115 device}  \phantom{000000} \\
	\multicolumn{7}{c}{ $\star$ refers to a two-stage (LPU+HPU) cascade architecture generated by \textit{CascadeCNN} \phantom{0} } \\
	\multicolumn{7}{c}{ 1. Expressed in: [GOp/s $\mathbf{(}$ GOp/s/DSP $\mathbf{)}$] using 18$\times$18 and 25$\times$18 DSP configurations } \\
	\multicolumn{7}{c}{ 2. Normalised performance over the 16-bit HPU of each network (in GOp/s/DSP) \phantom{000} } \\
\end{tabular}
}
\label{table:otherworks}
\vspace{-6mm}
\end{table}

\vspace{-2mm}
\section{Conclusion and Future Work}
\vspace{-1mm}
\textit{CascadeCNN}, an automated toolflow for CNN inference acceleration introducing a quantisation scheme to exploit the computation time-accuracy trade-off of CNNs, is presented. By designing a cascaded two-stage architecture, tailored to the target CNN-FPGA pair and user-specified error tolerance, the proposed framework demonstrates remarkable speed-up on both VGG-16 and AlexNet,
compared to a single-stage architecture for the same resource budget and accuracy. By exposing the application-level error tolerance to the design space exploration to increase the overall system throughput,
\textit{CascadeCNN} 
outperforms the current state-of-the-art FPGA-based CNN accelerators. Future work will focus on enhancing the tool with a latency-driven implementation of the cascade model, which would replace reconfiguration with resource sharing between the LPU and HPU. 

\vspace{-2.5mm}


\vspace{1.5mm}
{ \footnotesize 
\section*{Acknowledgment}
\vspace{-0.05cm}
The support of the EPSRC Centre for Doctoral Training in High Performance Embedded and Distributed Systems (HiPEDS, Grant Reference EP/L016796/1) is gratefully acknowledged. This work is also supported by EPSRC grant 1507723. 
}


%
\vspace{-0.3cm}
\bibliographystyle{IEEEtran}
\bibliography{main}

\end{document}